\documentclass[10pt,twocolumn,letterpaper]{article}

 \usepackage{cvpr}              

\definecolor{cvprblue}{rgb}{0.21,0.49,0.74}
\usepackage[pagebackref,breaklinks,colorlinks,allcolors=cvprblue]{hyperref}
\usepackage{amsmath, amssymb}
\usepackage{tikz}
\usetikzlibrary{positioning, shapes.geometric, arrows.meta, fit}
\def\paperID{*****} 
\def\confName{CVPR}
\def\confYear{2025}
\usepackage[accsupp]{axessibility}
\usepackage{orcidlink}

\usepackage{float}
\usepackage{placeins}
\usepackage{dirtytalk}
\usepackage{multirow}

\newtheorem{proposition}{Proposition}[section]
\newtheorem{proof}{Proof}[section]

\title{Confidence-calibrated covariate shift correction for few-shot classification in Vision-Language Models}

\author{
Behraj Khan\textsuperscript{1,3}\orcidlink{0000-0003-0985-9543},\ 
Rizwan Qureshi\textsuperscript{2}\orcidlink{0000-0002-0039-982X},\ 
Nouman Muhammad Durrani\textsuperscript{3}\orcidlink{0000-0001-6135-3924},\ 
Tahir Qasim Syed\textsuperscript{1*}\orcidlink{0000-0003-0638-9689} \\
\textsuperscript{1}School of Mathematics and Computer Science, Institute of Business Administration Karachi, Pakistan. \\
\textsuperscript{2}Center for Research in Computer Vision, University of Central Florida, USA. \\
\textsuperscript{3}National University of Computer and Emerging Sciences, Karachi, Pakistan. \\
{\tt\small \{behrajkhan, tqsyed\}@iba.edu.pk,   tahirqsyed@gmail.com}
}

\begin{document}
\maketitle
\maketitle
\thispagestyle{empty}

\begingroup
\renewcommand\thefootnote{}\footnote{* Corresponding author}
\addtocounter{footnote}{-1}
\endgroup

\begin{abstract}
Since the establishment of vision-language foundation models as the new mainstay in low-shot vision classification tasks, the question of domain generalization arising from insufficient target data is assuming more  importance. This scarcity challenge induces sampling bias and amplifies model sensitivity to variations and shifts in data distributions. While fine-tuning on multiple domains could mitigate such domain generalization issues, it is resource-intensive and demands diverse data sources. In this work, we systematically analyze two critical challenges: (1) covariate shift between the pre-training distribution and the underspecified target distribution, and (2) confidence misalignment, where predictions on novel data are overconfident. To address both challenges simultaneously, we introduce Confidence-Calibrated Covariate Shift Correction (CalShift)—a unified approach that combines a Fisher information penalty to mitigate covariate shift and a Confidence Misalignment Penalty (CMP) to reduce overconfidence in misclassified examples. Experimental evaluations across various vision and covariate shift benchmarks demonstrate that CalShift significantly improves model calibration, achieving up to a 5.82\% reduction in Expected Calibration Error (ECE). Furthermore, CalShift enhances robustness, improving accuracy by 3.5\% on challenging datasets impacted by covariate shifts. Our results highlight CalShift as a promising strategy for building robust and reliable low-shot vision-language systems for real-world applications.

\end{abstract}

\section{Introduction}
\label{sec: intr}

\begin{figure*}[htbp]
    \centering
    \begin{subfigure}[b]{0.5\textwidth}
        \centering
        \includegraphics[width=\textwidth]{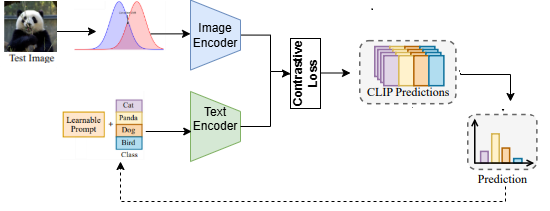}
        \caption{Misaligned predictions caused by covariate shift.}
        \label{fig:image1}
    \end{subfigure}
    
    \vspace{1 cm} 
    
    \begin{subfigure}[b]{0.52\textwidth}
        \centering
        \includegraphics[width=\textwidth]{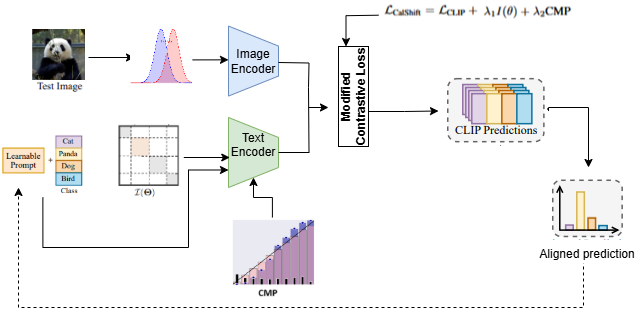}
        \caption{Aligned predictions after integration of CMP and FIM \(I(\theta)\).}
        \label{fig:image2}
    \end{subfigure}
    
    \caption{Workflow of the proposed CalShift framework: The sub-figure (a) illustrates the  confidence misalignment problem caused by covariate shift. The (bottom right) part of the sub-figure (a) show misaligned predictions. The subfigure (b) middle section represents the two components as recipe in method: Fisher Information Penalty (\(I(\theta)\)) for covariate shift correction and Confidence Misalignment Penalty (CMP) for calibration. Both FIM and CMP are integrated into  the CLIP  text encoder. The final loss, \underline{$\mathcal{L}_{\text{CalShift}} = \mathcal{L}_{\text{CLIP}} + \lambda_1 I(\theta) + \lambda_2 \text{CMP}$}, combines the original CLIP loss with FIM and CMP to produce robust and aligned predictions, as shown in (bottom right) as output.}
    \label{fig:combined}
\end{figure*}

Foundation models such as CLIP \cite{radford2021learning}, Flamingo \cite{alayrac2022flamingo}, and Align \cite{abdul2024align} form the cornerstone of contemporary vision-language understanding systems, especially for downstream vision classification tasks in the emerging post-training paradigm. Their  real-world deployment, while gaining ubiquity, has to contend with two significant challenges related to \textit{domain generalization}, the ability to generalize to classification tasks close to but outside the pre-training distribution.

The first challenge is \textit{covariate shift}, which occurs because the \textit{feature distribution} between training and target data differs, \(P(x)_{tr} \neq P(x)_{ts}\) while the conditional distribution remains unchanged \(P(y|x)_{tr} = P(y|x)_{ts}\) 
\cite{shimodaira2000improving,sugiyama2005input,cortes2014domain,khan2024causal}. This phenomenon is inherent in domain generalization for foundation models because feature distributions differ by definition, and is garnering attention as a significant angle of the generalization regime \cite{xiao2024any, murugesan2025robust, wen2024mitigating}. 

The second challenge, \textit{confidence misalignment}, emerges when a model's predicted confidence (softmax) scores fail to accurately represent the true predictive likelihood because of mis-classifications \cite{gal2016dropout,guo2017calibration, kumar2019verified}.   By definition, confidence calibration is the alignment between predicted probabilities and the true correctness likelihood \cite{wangopen, oh2023blackvip}.     This issue is particularly pronounced in   few-shot post-training     regimes, characterized by scarce labeled data, statistically unstable target distributions, and resulting unreliable predictions that are inadequately reflected in predictive confidence \cite{khan2025technical}.      Confidence misalignment mirrors harmful forgetting in LLMs \cite{huang2024vaccine}   when distribution shifts causes models uncertainty estimation to degrade. Similarly in our case, covariate shift can break the model confidence in VLMs, leading to overconfident errors.

Previous approaches such as \cite{khan2024causal} addresse the covariate shift problem caused by dataset fragmentation by regularizing the loss function with Fisher information penalty. \cite{murugesan2025robust} highlight calibration degradation of CLIP-based adaptation methods in presence of distribution shifts particularly in out-of-distribution (OOD) scenarios due to increased logit ranges. The authors propose a model-agnostic logit scaling approach to restore calibration while maintaining performance.  \cite{wangopen} addresses the problem of miscalibration in vision-language models and the author propose data-based approach such as, dynamic outlier regularization (DOR). DOR mitigates the calibration problem by minimizing feature deviation for novel textual labels to balance calibration between new and base classes. \cite{khan2025technical} study the confidence calibration problem under covriate shift in low-shot setting for vision-language model. The authors resolve this problem using confidence alignment. 

These approaches  typically address either covariate shift or confidence misalignment independently. To fill this gap, we propose a unified framework, termed \textit{Confidence-  Calibrated Covariate Shift Correction (CalShift)}, which jointly addresses both challenges. Our method employs the popular Tikhonov regularization principle. It therefore integrates a Fisher information-based \cite{lehmann2006theory}  penalty to mitigate covariate shift and a confidence misalignment penalty (CMP) to recalibrate confidence scores within the contrastive
loss formulation of CLIP.      CMP penalizes overconfident incorrect predictions by shifting log-likelihood to the true class. In doing so, CalShift provides a robust mechanism for achieving reliable and well-calibrated predictions in low-shot vision classification scenarios. The working mechanism of our framework is shown in Figure \ref{fig:combined}.

Furthermore, our method CalShift aligns with growing interest in training-free adaptation, where a models adapts to new or unseen task without having access to any additional data at training time. Our methods allows VLMs to adapt to new target distribution in a training-free manner, making it highly effective in real-world applications where access to more training data is costly or infeasible.

\subsection*{Contributions}
In this work, we bridge two distinct yet related areas of research by unifying penalties previously studied independently for covariate shift and confidence misalignment. 

  
\begin{enumerate}
    \item We introduce a loss component that suppresses CLIP's confidence on misclassifications by moderating the flow of posterior likelihood towards any posterior peak higher than that of the true class.
    
    \item We use knowledge about incorrect predictions existing  as labels, which is inherently invariant to transformations such as data augmentation or batch/layer-normalization. 
    
    \item  We discover that correcting for  covariate shift in relaxed conditions of $P(x|y)$ not necessarily staying constant, still helps improves calibration in real-world settings.   
    \item We offer improvements in the predictive envelope of a post-trained model along the axes of confidene  calibration, trustworthiness and generalisation. 
\end{enumerate}

\section{Related work}
\textbf{Vision-language foundation models.} Foundation models open new horizons for machine learning tasks, particularly in supervised learning \cite{radford2021learning, jia2021scaling}. Foundation models like CLIP can be easily adapted for downstream tasks. To further improve, downstream task classification, several efficient methods are available, including parameter-efficient prompt tuning (PEFT) \cite{zhou2022learning}, prompt tuning, vanilla fine-tuning \cite{khattak2023maple}, adapter tuning \cite{zhang2021tip}, and CLIP fine-tuning \cite{ gao2024clip}. 

The CLIP foundation model  \cite{radford2021learning}, consist of an image encoder \(f(x)\) and a text encoder \(g(\textbf{t})\). Both encoders are jointly trained using contrastive loss. The pre-trained CLIP provide support of few-shot learning which can then be used for downstream tasks like \textit{prompt learning} using fixed or hand-crafted prompts.
Context Optimization (CoOp) \cite{zhou2022learning} introduced learnable task-specific prompts by replacing the manually designed prompts. CoOp provides two different prompt implementation i.e. \textit{Unified context vectors} where prompts shared across classes and \textit{Class-specific context vectors} provides individual prompts per class. But admittedly,  CoOp struggles with generalization to unseen classes and distribution shift.

\noindent
\textbf{Distribution shift in  foundation models.}Distribution shifts present unique challenges for foundation models compared to traditional domain generalization or adaptation \cite{hendrycks2019robustness,wortsman2022robust}. In classical approaches, models are trained to learn invariant features under the assumption that these invariances hold at test time. In contrast, foundation models like CLIP \cite{radford2021learning}, ALIGN \cite{jia2021scaling}, and Flamingo \cite{alayrac2022flamingo} are pre-trained on large-scale data and later adapted to downstream tasks. However, deploying these models in low-shot settings introduces challenges such as covariate shift (where the input distribution \(P(x)\) differs between pretraining (source) and target data) and confidence misalignment (where the model becomes overconfident on shifted inputs) \cite{ wang2023calibration, wang2024understanding,khan2025technical}. \cite{huang2024vaccine} highlighted how distribution shifts can induce confidence misalignment in VLMs and proposed \say{vaccine} approach, applying perturbation-aware confidence alignment during fine-tuning to mitigate harmful embedding shifts. We address confidence alignment in low-shot settings through calibration-aware regularization.

\noindent\textbf{Confidence calibration in deep learning models.}
Extensive research focuses on confidence calibration methods to ensure model accuracy is aligned with its predicted confidence scores. These methods includes regularization based approaches, such as implicit regularization \cite{ross2017focal} \cite{baloch2019focused}, \(L_{2}\) regularization \cite{guo2017calibration} , and entropy regularization \cite{pereyra2017regularizing} to align the model predicted accuracy with confidence. Some augmentation-based approaches, such as label-smoothing \cite{muller2019does} and mix-up training methods \cite{thulasidasan2019mixup, zhang2022and} are also explored for confidence calibration in deep learning models. 
\cite{ovadia2019can} empirically investigated the calibration of deep learning models under distribution shift. A comprehensive survey \cite{wang2023calibration}, which addresses the recent development about calibration for deep learning models.

\noindent\textbf{Confidence calibration in foundation models.}
\cite{pandeyconfident} addresses the issue of under-confidence in pre-trained foundation models when these models are fine-tuned for downstream tasks in few-shot settings. Similarly, \cite{murugesan2025robust} highlights the miscalibration problem in CLIP-based adaptive approaches. Their proposed method empirically demonstrates that CLIP-based adaptive approaches, such as prompt learning, adapters, parameter-efficient fine-tuning (PEFT), and test-time adaptation, are prone to miscalibration in low-shot settings under distribution shifts. Furthermore, \cite{tu2024toward} investigates the prediction uncertainty of the foundation model CLIP for image classification tasks, focusing on variations in visual factors. Recent work \cite{wangopen} observed observed the post-hoc confidence miscalibration in fine-tuned VLMs and proposed Distance-Aware Calibration (DAC) to align calibration of VLMs.

In this work, we propose a regularization-based approach to ensure that the predictive probabilities of foundation models align with confidence scores in a prompt-learning setting under distribution shifts. 
\section{    Method    }
\label{sec: mthd}

Contrastive Language–Image Pretraining (CLIP) is a multimodal model developed by \cite{radford2021learning} that learns latent-space image-text association from large datasets and enables zero-shot transfer learning to diverse vision tasks.  CLIP uses contrastive loss (\(\mathcal{L}_{\text{c}}\) for image-text alignment can be defined as: \[
\label{clip:loss}
  \mathcal{L}_{\text{c}} = \frac{1}{2} \left( \mathcal{L}_{\text{txt}} + \mathcal{L}_{\text{img}} \right),  
\]
while, text loss is given as: \[
\label{clip:txtloss}
 \mathcal{L}_{\text{txt}} = -\frac{1}{N} \sum_{i=1}^N \log \frac{\exp\left(\text{sim}(\mathbf{t}_i, \mathbf{i}_i) / \tau\right)}{\sum_{j=1}^N \exp\left(\text{sim}(\mathbf{t}_i, \mathbf{i}_j) / \tau\right)},   
\]
and image loss is denoted as:  \[
\label{clip:imgloss}
 \mathcal{L}_{\text{img}} = -\frac{1}{N} \sum_{i=1}^N \log \frac{\exp\left(\text{sim}(\mathbf{i}_i, \mathbf{t}_i) / \tau\right)}{\sum_{j=1}^N \exp\left(\text{sim}(\mathbf{i}_i, \mathbf{t}_j) / \tau\right)}   
\]. 

    \noindent\textbf{Covariate shift:} In low-shot learning regimes, availability of limited data may be insufficient to bring the post-training distribution simulate the pre-training distribution, which leads to poor generalization. Similarly, when data are distributed across batches or folds, each subset represents an extremely low-data setting, where distribution shift emerges due to insufficient sample representation \textit{between} any two subsets of post-training data. The information loss due to covariate shift caused by such fragmented data partitions \cite{sugiyama2007covariate, moreno2012study} can be effectively measured by Fisher information \cite{khan2024causal,vigneron2010adaptive, courtade2016monotonicity}, very similar to in low-shot learning scenarios in classical transfer learning.\\

\noindent\textbf{Fisher information in covariate shift}. 
Fisher Information has been shown to an equivalent of relative entropy without iterative computation, and accesses the curvature of the parameter surface to measure even small perturbatiion
\cite{lehmann2006theory, martin2016lecture}.  Fisher information measures the sensitivity of model to change in data distribution makes it ideal solution to diagnose and mitigate shift in covariates in VLMs. The Fisher information measure can be defined as:
    
 \begin{equation}
 \label{eq:FiM}
 I(\theta) = -\mathbb{E}\left[\frac{\partial^2 \log P(X;\theta)}{\partial \theta^2}\right]   
\end{equation}
    
 where \(I(\theta)\) is Fihser information which can be defined as the negative hessian of the log-likelihood function \(-\mathbb{E}\left[\frac{\partial^2 \log P(X;\theta)}{\partial \theta^2}\right] \) \cite{lehmann2006theory,martin2016lecture}.

 Recently, the FI matrix was used to compute and by consequence, help correct, divergence in data distributions under a batched-streaming setting \cite{khan2024causal}. We are possibly the first to  borrow the idea into the VLM post-training regime.\\

\noindent\textbf{Confidence misalignment: }     Foundation models such as CLIP are often overconfident in fine-tuned predictions when image-text pair does not reconcile, specifically in low-shot settings \cite{tu2024toward, pandeyconfident, murugesan2025robust, khan2025technical}. The   CLIP contrastive loss maximizes \(\mathcal{L}_c\) for correct image-text pair while minimizing it for incorrect pairs.  Evidence is emerging of \textbf{neural collapse} in the final phase of training/fine-tuning/post-training\cite{papyan2020prevalence}. It is noticed that intra-class examples close in or collapse towards the class mean. The rescaling by the softmax to $1$ may further reduces  numerical variability. The higher and narrower peak encourage over-confidence, regardless of the correctness of the classification,
which results in  \(H(P(\mathbf{t}|\mathbf{i}; \theta)) \rightarrow 0\). \\

\noindent\textbf{Fisher information and CMP.} Fisher information addresses the distribution shift and uncertainty caused by covariate shift while CMP helps in reliability and calibration of model. Both challenges are closely linked: As covariate shift exacerbates confidence misalignment by creating distributional discrepancies, whereas overconfident predictions can obscure the underlying uncertainty caused by distribution shifts \cite{khan2024causal,murugesan2025robust}. 

To overcome the overconfidence problem, the confidence misalignment penalty (CMP) is introduced by \cite{khan2025technical}, which draws likelihood away from a text-image-pair mismatch. This likelihood is moved into  the  matching class from each mismatched class pair in proportion to the mismatched likelihood. That excess likelihood is:
\begin{equation}
\label{eq:cmp}
\text{CMP} = \frac{P(x, y)}{\sum_{\substack{y' \neq y \ P(x, y') > P(x, y)}} P(x, y')},
\end{equation}

\noindent where \(P(x,y)\) is the softmax probability for the correct pair \text{(}i, t\text{)},  while \(P(x, y')\) is the softmax probability for incorrect pairs \text{(}i, t'\text{)}, where \(t \neq t'\). 


CalShift which combines Fisher information and CMP into a single cohesive approach    and jointly regularize      the CLIP contrastive loss,  This holistic approach provides a non-post-hoc  way for confidence calibration under covariate shift of VLMs in few-shot learning regimes.  CalShift controls misalignment, thereby enhancing VLM reliability under distribution shifts and reducing harmful overconfidence in safety-critical applications.

    
\begin{proposition}
    The Fisher information \(I(\theta)\) improves generalization by controlling the curvature of the loss landscape and CMP improves calibration by penalizing overconfidence and redistributing log-likelihood to the true class.  
\end{proposition}

    The Fisher information \(I(\theta)\) in eq \ref{eq:FiM} measures the curvature of log-likelihood function with respect to model parameters \(\theta\). A high \( I(\theta) \) indicates sharp curvature which results in poor generalization. To mitigate this effect, we introduce Fisher information regularization by adding \( \lambda_1 I(\theta) \) into the loss function: 
    \[\mathcal{L}_{\text{CalShift}} = \mathcal{L}_{\text{c}} + \lambda_1 I(\theta).\]
From PAC-Bayes perspective \cite{wang2021pac,hellstrom2025generalization}, \( I(\theta) \) tightens the generalization bound by controlling the complexity of hypothesis class. The PAC-Bayes bound for a model with parameters \(\theta\) is:
\[\mathcal{E}_{\text{gen}}(\theta) \leq \mathcal{E}_{\text{emp}}(\theta) + \frac{I(\theta) + \log(1/\delta)}{2n}\]

where, \(\mathcal{E}_{\text{gen}}(\theta)\) is generalization error and can be defined as: 
\[\mathcal{E}_{\text{gen}}(\theta) = \mathbb{E}_{(x,y) \sim P_{\text{test}}} [\ell(f_{\theta}(x), y)],\] 
while \(\mathcal{E}_{\text{emp}}(\theta)\) is empirical error and is given by:
\[\mathcal{E}_{\text{emp}}(\theta) = \frac{1}{n} \sum_{i=1}^{n} \ell(f_{\theta}(x_i), y_i),\]
\(\theta\) is confidence parameter and \(n\) is number of training samples. Minimization of \( I(\theta) \),  reduce \(\mathcal{E}_{\text{gen}}(\theta)\), ensures model robustness to distribution shift.

The second term, CMP given in eq \ref{eq:cmp} penalizes overconfidence by redistributing log-likelihood to the true class, it ensures that does not assign excessively high probabilities to incorrect classes. 
To mitigate overconfidence we integrate \( \lambda_2 CMP \) into loss function:\[\mathcal{L}_{\text{CalShift}} = \mathcal{L}_{\text{c}} + \lambda_2 CMP.\]
From a Bayesian perspective, CMP act as regularization that encourage the model to distribute probability mass more uniformly across plausible classes. This helps in model calibration by aligning the predicted probabilities with actual correctness likelihood.

From the Minimum Description Length (MDL) principle \cite{hansen2001model,kalai2024calibrated,zhao2024large}, the CMP reduces the complexity of learned hypothesis by controlling overconfident predictions. The MDL bound for  a model with \( \theta \) parameters can be defined as: 

\[\mathcal{E}_{\text{gen}}(\theta) \leq \mathcal{E}_{\text{emp}}(\theta) + \frac{\mathcal{C}(\theta) + \log(1/\delta)}{2n}.\]
Where \(\mathcal{C}(\theta)\) is hypothesis complexity. By minimizing CMP, model effectively reduces the hypothesis complexity, provides a more compact representation that generalizes better. It helps the model to form robust decision boundaries.

\( I(\theta) \) regularization ensures that representations remain robust under domain shifts and CMP ensures that the model does not become overconfident, improving calibration.  Since neither penalty degrades the other’s effect, CalShift optimally balances robustness and calibration.

\begin{table*}[!ht]
\centering
\caption{Few-shot accuracy (\%) on ImageNet with CalShift. The $\Delta$ row in upper half of the table shows improvement (\textbf{$\uparrow$}) or degradation (\textbf{$\downarrow$}) in accuracy over baseline. The bottom half of the table shows Expected Calibration Error (ECE \%) on ImageNet with and without CMP penalty. In ECE scenario lower is better. $\Delta$ shows percentage improvement (\textbf{$\downarrow$}) or degradation (\textbf{$\uparrow$}).}
\label{tab:fewshot_results}

\begin{tabular}{llccccccc} 
\toprule
                          \multicolumn{9}{c}{\textbf{~~~~~~~~~~~~~~~~~~~~~~~~~~~~~~~~~~~~~Number of shots}}                 \\ 
\cmidrule{3-9}
\multirow{6}{*}{\textbf{ACC }} & \textbf{Method} & \textbf{0} & \textbf{1} & \textbf{2} & \textbf{4} & \textbf{8} & \textbf{16} & \textbf{Avg.} \\ 
\cmidrule{2-9}
                               & CLIP            & 72.4 & 68.1 & 70.3 & 73.6 & 75.9 & 77.2 & 72.9 \\ 
\cline{2-9}
                               & CoOp            & 79.5 & 75.2 & 77.8 & 80.1 & 82.4 & 83.7 & 79.8 \\ 
                               & CoOp + FIM      & \textbf{84.9} & \textbf{80.6} & \textbf{82.1} & \textbf{84.4} & \textbf{86.7} & \textbf{88.0} & \textbf{84.5} \\ 
\cline{2-9}
                              & $\Delta$ \%     & \textbf{6.8 ↑} & \textbf{7.2 ↑} & \textbf{5.5 ↑} & \textbf{5.4 ↑} & \textbf{5.2 ↑} & \textbf{5.1 ↑} & \textbf{5.9 ↑} \\
\bottomrule
\\
\bottomrule

     \multirow{6}{*}{\textbf{ECE }}                           & CLIP            & 1.51 & 2.89 & 2.67 & 2.12 & 1.98 & 1.75 & 2.15 \\ 
\cline{2-9}
                               & CoOp            & 3.36 & 3.12 & 2.94 & 2.88 & 2.64 & 2.45 & 2.90 \\ 
                               & CoOp + CMP      & \textbf{3.06} & \textbf{2.84} & \textbf{2.71} & \textbf{2.52} & \textbf{2.38} & \textbf{2.20} & \textbf{2.62} \\ 
\cline{2-9}
                              & $\Delta$ \%     & \textbf{8.93 ↓} & \textbf{8.97 ↓} & \textbf{7.82 ↓} & \textbf{12.50 ↓} & \textbf{9.85 ↓} & \textbf{10.20 ↓} & \textbf{9.66 ↓} \\
\bottomrule
\end{tabular}
\end{table*}

 
\begin{table*}[!ht]
\centering
\caption{The upper half shows CalShift accuracy results on vision datasets with and without FIM penalty. The $\Delta$ row shows the percentage increase (\textbf{$\uparrow$}) or decrease (\textbf{$\downarrow$}) in accuracy. The lower half shows CalShift ECE results on vision datasets with and without CMP penalty. The $\Delta$ row shows the percentage increase (\textbf{$\uparrow$}) decrease (\textbf{$\downarrow$})  in calibration}
\label{tab:acc_results}
\resizebox{\textwidth}{!}{%
\begin{tabular}{llcccccccccccc} 
\toprule
\multirow{6}{*}{\textbf{ECE }} & \textbf{Method} & \textbf{UCF101}      & \textbf{Food101}     & \textbf{Caltech101}  & \textbf{OxfordPets}        & \textbf{Flowers102}  & \textbf{ImageNet}    & \textbf{StanfordCars}        & \textbf{FGVCAircraft}    & \textbf{SUN397}      & \textbf{DTD}         & \textbf{EuroSAT}     & \textbf{Avg.} 
       \\ 
\cmidrule{2-14}
                               & CLIP            & 69.9  & 90.1 & 96.8 & 91.2 & 72.1 & 72.4 &  63.3 & 27.2 & 69.4 & 53.3 & 56.5                    &   69.3              \\ 

\cline{2-14}
                               & CoOp            & 78.6 & 97.0 & 98.6 & 98.2 & 79.2 & 79.5 & 59.2 & 25.2 & 63.0 & 52.5 & 53.8                      &  71.2                     \\ 
                               & CoOp + FIM        & \textbf{84.3} & \textbf{98.7} & \textbf{98.8} & \textbf{98.9} & \textbf{85.5 }& \textbf{84.9} & 54.3 & \textbf{27.2} & \textbf{66.8} & \textbf{55.1} & 49.2                      &  \textbf{73.5}                \\ 
                                \cline{2-14}
                              & $\Delta$ \%     & \textbf{7.3  ↑} & \textbf{1.7  ↑} & \textbf{0.2  ↑} & \textbf{0.7  ↑} & \textbf{8.0  ↑} & \textbf{6.8  ↑} & \textbf{8.2  ↓} & \textbf{7.9  ↑} & \textbf{6.0  ↑} & \textbf{4.9  ↑} & \textbf{8.6  ↓} & \textbf{3.2  ↑}  \\
\cmidrule{2-14}
\\
\multirow{6}{*}{\textbf{ACC }} &  &      &     &   &        &   &    &         &    &      &         &     &           \\ 
\cmidrule{2-14}
                               & CLIP            &     3.24             &   1.57               &    6.49              &    2.25             &       3.11          &    1.51             & 3.74                 &   3.03               &    1.59              &       4.53          &       9.12           &      3.52             \\ 

\cline{2-14}
                               & CoOp            & 3.08                & 3.35          & 3.24              & 3.06         & 2.96            & 3.36             & 3.38          & 3.24            & 3.02             & 3.06              & 3.08                & 3.16                      \\ 
                               & CoOp + CMP       &\textbf{ 2.94}                & \textbf{3.02}          & \textbf{3.08}              & \textbf{2.84}         & 3.16            & \textbf{3.06}             & \textbf{3.16}          & \textbf{3.08}            & \textbf{2.96}             & \textbf{2.92}              & \textbf{3.06}                & \textbf{2.98}                        \\ 
                                \cline{2-14}
                              & $\Delta$ \% & \textbf{4.55}$\downarrow$  & \textbf{9.85}$\downarrow$ & \textbf{4.94}$\downarrow$  & \textbf{7.19}$\downarrow$  & 6.76$\uparrow$  & \textbf{8.93}$\downarrow$  & \textbf{6.51}$\downarrow$  & \textbf{4.94}$\downarrow$  & \textbf{1.99}$\downarrow$  & \textbf{4.57}$\downarrow$ & \textbf{0.65}$\downarrow$ & \textbf{5.70}$\downarrow$                      \\
\bottomrule
\end{tabular}}
\end{table*}
\noindent
Thus, minimizing the loss function given in eq \ref{eq: CalShiftloss}, guarantees a pareto-optimal balance between robustness to covariate shift and also preserving the confidence calibration.
\begin{equation}
\label{eq: CalShiftloss}
\mathcal{L}_\text{CalShift}(x,y; \theta) = \mathcal{L}_c + \lambda_1 I(\theta) + \lambda_2 \text{CMP} 
\end{equation}

More detail on the theoretical background about \(I(\theta)\) and CMP can be found in   Appendix \ref{sec: tj}.
\section{Experiments and Results}

\subsection{Experiments}
\label{sec:exp-and-reslt}
\noindent\textbf{Datasets.} To check the effectiveness of CalShift, we used 19 vision dataset including natural distribution datasets, domain adaptation datasets and vision classification datasets used for evaluation by low-shot prompt learning methods. These datasets are:  {Caltech101}  \cite{FeiFei2004LearningGV}, Imagenet \cite{deng2009imagenet}, EuroSAT \cite{helber2019eurosat}, StanfordCars \cite{krause20133d}, FGVCAircraft \cite{maji2013fine}, OxfordPets \cite{parkhi2012cats}, SUN397 \cite{xiao2010sun}, Food101 \cite{bossard2014food}, Flowers102 \cite{nilsback2008automated}, DTD \cite{cimpoi2014describing}, UCF101 \cite{soomro2012ucf101}, PACS \cite{li2017deeper}, VLCS \cite{fang2013video}, Office-Home \cite{venkataraman2016sparkr} and five domain adaptation datasets such as Imagent-1k and its variants like ImageNet A (Art) \cite{hendrycks2021natural}, ImageNet-V2 (V2) \cite{recht2019imagenet}, ImageNet R (Real) \cite{hendrycks2021many} and ImageNet S (Sketch) \cite{wang2019learning}.\\

\noindent\textbf{Baselines.}
We consider CLIP and prompt learning method CoOp\cite{zhou2022learning} without any penalty integration as baseline in comparison to our method.\\


\noindent\textbf{Evaluation metrics.} We used accuracy (Acc) as evaluation metric for robustness of CalShift under covariate shift and vision datasets used by prompt learning approaches, while Expected Calibration Error (ECE) \cite{guo2017calibration} is used with same number of datasets for evaluating the calibration performance of CalShift.\\  

\noindent\textbf{Implementation details.} To check the robustness and calibration performance of CalShift, we conduct extensive experiments on various datasets. We used pre-trained CLIP \cite{radford2021learning} with ViT-B/16 \cite{dosovitskiy2020image} as backbone in prompt learning method. In covariate shift  experiment we used CLIP with ViT-B/16 and ResNet-50 backbone \cite{he2016deep} for the last four variants of ImageNet. For first four datasets we follow the same protocol of \cite{li2017deeper}, where model is trained on another domain and evaluated on another test domain.

 The experimental detail, datasets, and evaluation metrics used in the paper can be found in appendix \ref{sec:exp-and-reslt}. here,

we discuss the robustness and confidence calibration results of our method CalShift.
 \subsection{Results}
     
\noindent\textbf{Calshift generalization performance across shots.} Table \ref{tab:fewshot_results} upper part shows the generalization performance of CalShift across incremental few-shot settings. It is shown in Table \ref{tab:fewshot_results} that the integration of FIM with CoOp consistently outperforms CLIP and vanilla CoOp across all shot settings. CalShift achieves a maximum accuracy improvement of $6.8\%$  over CoOp in the zero-shot setting and $7.2\%$ in the 1-shot setting. CalShift shows a $5.9\%$ average accuracy improvement over CoOp, demonstrating that CalShift regularization effectively mitigates overfitting in data-scarce regimes. 

The $\Delta \%$ decrease monotonically from $7.2\%$ ($1$-shot) to $5.1\%$ ($16$-shot) settings, which shows the effectiveness of CalShift in extremely low-shot data regime. This provides empirical strength to FIM's role in stabilizing gradient updates when labeled examples are scarce. Furthermore, in a zero-shot setting with no training data, integrating FIM with CoOp surpasses CLIP, demonstrating FIM's superior ability to retain knowledge compared to CoOp alone. This suggests that FIM acts as a prior-preserving regularizer, preventing CLIP from catastrophic forgetting in zero-shot scenarios.
\\

\noindent\textbf{Calshift calibration performance across shots.}
CalShift's calibration performance results are shown in the lower half of Table \ref{tab:fewshot_results}. Integrating CMP with CoOp consistently outperforms vanilla CoOp across all few-shot settings, achieving up to a $12.5\%$ improvement in the $4$-shot setting and $10.2\%$ in the $16$-shot scenario. In the zero-shot scenario, CLIP's calibration performance surpasses CoOp, indicating that prompt tuning can degrade CLIP's calibration. Integrating CMP into CoOp mitigates this issue, improving calibration by $8.93\%$ in the zero-shot setting.

Integrating CMP into CoOp improves $\Delta\%$ by a maximum of $8.97\%$ in the $1$-shot setting and $10.2$\% in the 16-shot setting, demonstrating CMP's ability to better leverage additional data for confidence calibration. Table \ref{tab:fewshot_results} demonstrates that integrating CMP into CoOp outperforms both vanilla CoOp and CLIP beyond the 4-shot setting, suggesting that CMP enhances calibration adaptation. This suggests that CMP tackles CoOp’s two key issues by penalizing overconfident logits: (1) overfitting in few-shot scenarios and (2) gradient misalignment . CMP addresses overfitting by enforcing decision boundary constraints, preventing overconfidence on limited data. Meanwhile, it mitigates misalignment by aligning logit gradients with true class margins, reducing bias in probability estimates.

Table \ref{tab:fewshot_results} shows that FIM integration significantly enhances generalization, particularly in low-shot regimes, while CMP integration improves calibration in these settings. These results establish CalShift as a versatile regularizer, ensuring robust and trustworthy adaptation in vision-language models.\\

\begin{table*}[!ht]
\centering
\caption{The upper part shows CalShift accuracy results on covariate shift vision datasets with and without FIM penalty. The $\Delta$ row shows the percentage increase/decrease in accuracy. \textbf{$\uparrow$} shows improvement in accuracy while the \textbf{$\downarrow$} shows decrease in performance. The lower part shows CalShift ECE results on covariate shift vision datasets with and without CMP penalty. The $\Delta$ row shows the percentage increase/decrease in calibration. \textbf{$\downarrow$}  shows improvement in calibration while \textbf{$\uparrow$} shows calibration performance decrease.}
\label{tab:acc_cvt_results}
\resizebox{\textwidth}{!}{%
\begin{tabular}{llcccccccc} 
\toprule
\multirow{6}{*}{\textbf{ACC }} & \textbf{Method} & \textbf{PACS} & \textbf{Office-Home}        & \textbf{VLCS}         & \textbf{DomainNet}   & \textbf{ImageNet-V2}  & \textbf{ImageNet-S}  & \textbf{ImageNet-A}  & \textbf{ImageNet-R }  \\ 
\cmidrule{2-10}
                               & CLIP            &    96.1                  &  80.4                    &     81.4                &    54.1                  &    60.8                   &  46.2                    &    47.8                  &     73.9                  \\ 

\cline{2-10}
                               
                               & CoOp            &      96.5                &   82.1                   &    82.5                  &                    58.8  &        64.2               &      47.9                &      49.7                &    75.2                   \\ 
                               & CoOp+FIM       &      \textbf{98.0}                &   \textbf{85.6}                   &   \textbf{ 86.0}                  &      \textbf{60.0}                &        \textbf{65.5 }              &      \textbf{48.8}                &      \textbf{50.5}                &    \textbf{76.8}                   \\ 
\cline{2-10}
                          &  $\Delta$ \%& \textbf{1.5} $\uparrow $ & \textbf{3.5} $\uparrow $ & \textbf{3.5} $\uparrow $ & \textbf{1.2} $\uparrow $ & \textbf{1.3} $\uparrow $ & \textbf{0.9} $\uparrow $ & \textbf{0.8} $\uparrow $ & \textbf{1.6} $\uparrow $             \\ 
\cmidrule{2-10}
\\
\multirow{7}{*}{\textbf{ECE }} &  &  &         &        &    &   &   &   &   \\ 
\cmidrule{2-10}

                              & CLIP & 2.18 & 2.77 & 2.89 & 3.69 & 2.44 & 4.88 & 8.34 & 3.51 \\ 
\cmidrule{2-10}

                              & CoOp & 2.02 & 2.92 & 3.01 & 3.29 & 4.19 & 8.40  & 15.34 & 3.12 \\ 

                              & CoOp+CMP & \textbf{1.91} & \textbf{2.75} & \textbf{2.84} & \textbf{3.15} & \textbf{4.05} & \textbf{8.18} & \textbf{15.00} & \textbf{2.95} \\ 
\cmidrule{2-10}
                              & $\Delta$ \% & \textbf{5.45} $\downarrow$ & \textbf{5.82} $\downarrow$ & \textbf{5.64} $\downarrow$ & \textbf{4.26} $\downarrow$ & \textbf{3.34} $\downarrow$ & \textbf{2.62} $\downarrow$ & \textbf{2.21} $\downarrow$ & \textbf{5.45} $\downarrow$ \\ 
\bottomrule
\end{tabular}}
\end{table*}

\noindent\textbf{CalShift robustness in vision tasks.} The results in Table \ref{tab:acc_results} demonstrates the effectiveness of FIM penalty integration into CoOP as compared to baselines. The $\Delta$ row in Table \ref{tab:acc_results} shows improvement across most datasets. The notable improvement in accuracy include Imagenet ($6.8\%$), Flowers$102$ ($8.0\%$), and UCF$101$ ($7.3\%$), indicates significant improvement, from action recognition to fine-grained classification. In scene and texture-based datasets CalShift shows robustness  performance improvement with SUN$397$ ($6.0\%$) and DTD ($4.9\%$). For Food$101$ ($1.7\%$), Caltech$101$ ($0.2\%$), and OxfordPets ($0.7\%$)  improvement in accuracy. CalShift improves in accuracy by $3.2\%$ on average across all datasets.
\\

\noindent\textbf{CalShift calibration in vision tasks.} The CMP penalty-based calibration performance results across eleven vision datasets are shown in Table \ref{tab:acc_results}. In Table \ref{tab:acc_results}, it is shown that CMP integration improves in calibration for most datasets with an average of $5.70\%$ reduction in calibration. 

Smaller values are better in the calibration case. Penalty-based CoOp shows $9.85\%$ calibration improvement at maximum on Food101, mageNet ($8.93\%$), and OxfordPets ($7.19\%$), which demonstrates CMP alignment in model calibration in these settings. On other datasets also it shows better performance such as, UCF$101$ ($4.55\%$), Caltech101 ($4.94\%$), StanfordCars ($6.51\%$), and FGVCAircraft ($4.94\%$).

In comparison to CLIP, CoOP is surpassed by CLIP but when penalty is added then CoOp shows better calibration. Integration of penalty improves calibration across almost all datasets except Flowers$102$, where calibration is slightly worsen. The results provides an emprical evidence that penalty integration is reliable technique for reducing overconfidence and improving model reliability in low-shot learning.

Comparing penalty-bases CoOp to CLIP, penalty-based enhancement provides significant improvement across almost all datasets except StanfordCars and EuroSAT. CalShift shows UCF$101$ ($14.4\%$), Flowers$102$ ($13.4\%$), and ImageNet ($12.5\%$) improvement at maximum in comparison to CLIP.  

The integration of FIM penalty generally shows improvement in accuracy in low-shot vision tasks 
in particular when combined with prompt learning methods.\\

\noindent\textbf{CalShift robustness in covariate shift.} Table \ref{tab:acc_cvt_results} shows the robustness performance of CalShift on eight domain generalization datasets under covariate shift. The $\Delta$ of Table \ref{tab:acc_cvt_results} shows the penalty-based CoOp outperform CoOp across all datasets, showing significant improvement in robustness to covariate shift.  CalShift shows maxmimum robustness performance in Office-Home ($3.5\%$) and VLCS ($3.5\%$) demonstrating FIM integration enhances adaptation across domain adaptation tasks. For all other datasets such as, DomainNet ($1.2\%$), ImageNet-V2 ($1.3\%$), ImageNet-S ($0.9\%$), ImageNet-A ($0.8\%$), and ImageNet-R ($1.6\%$), CalShift shows slight better improvement, suggesting better robustness in challenging real-world distribution shift.

While comparing penalty-based CoOp to CLIP, we observe consistent improvement in accuracy across all domain generalization datasets like, VLCS ($4.6\%$) , Office-Home ($5.2\%$) and PACS ($1.9\%$). CalShift demonstrates better robustness on more challenging variants of ImageNet such as,  ImageNet-V2 ($1.3\%$) ,  ImageNet-S ($2.6\%$), ImageNet-A ($2.7\%$), and ImageNet-R ($2.9\%$). 

The FIM penalty indicates consistent performance improvement across covariate shift datasets, reinforcing the FIM penalty role in domain generalization and robustness to covariate shift datasets.   
\\

\noindent\textbf{CalShift calibration in covariate shift.} Table \ref{tab:acc_cvt_results} presents, calibration performance of penalty integration into CoOp for covariate shift vision datasets. It is demonstrated, that CalShift improves in calibration across all datasets with an average of $2.21\%$ at minimum and  $5.82\%$ at maximum. CalShift shows calibration improvement on Office-Home ($5.82\%$), PACS ($5.45\%$), VLCS ($5.64\%$), ImageNet-R ($5.45$\%) DomainNet ($4.26\%$) and ImageNet-V2 ($3.34\%$), while smaller but consistent improvements are seen in ImageNet-S ($2.62\%$) and ImageNet-A ($2.21\%$), which indicates the CMP effectiveness in confidence alignment on these datasets.

CLIP shows better calibration as compared to CoOP by reporting lower ECE such as, PACS ($2.18$ vs. $2.02$), VLCS ($2.89$ vs. $3.01$), and ImageNet-R ($3.51$ vs. $3.12$). However, when penalty is integrated into CoOp, the calibration reduces significantly making it more comparable to CLIP. \\

\noindent\textbf{Ablation of loss components.} We evaluate the performance of our proposed architecture by conducting ablation studies to understand the relative contribution of each component, Fisher information and CMP. We ablate the model by keeping the Fisher information \(I(\theta)\) hyperparameter \(\lambda_1 = 0\) in Equation \ref{eq: CalShiftloss} and then perform the experiment. Similarly, we set the CMP hyperparameter \(\lambda_2 = 0\) and conduct another experiment. The results are reported in Table \ref{tab:ablation_fim} and Table \ref{tab:ece_no_fim}, respectively.\\

\noindent\textbf{Effect of Fisher Information.} Table \ref{tab:ablation_fim} evaluates the isolated contributions of CMP and FIM across 11 vision datasets. The upper part of the table presents CalShift's generalization performance with 
$\lambda_1 = 0 $ and $\lambda_2 = 0.4$. CalShift demonstrates generalization efficacy, achieving a maximum improvement of $3.3\%$ on Flowers102, $1.9\%$ on SUN397, and $2.6\%$ on ImageNet. However, it shows only marginal accuracy gains on OxfordPets and Caltech101, suggesting that CMP's effect diminishes when CLIP features are already sufficient.

The lower part of Table \ref{tab:ablation_fim} presents CalShift's accuracy across 11 vision datasets with 
$\lambda_1 = 0.4 $ (FIM-only) and $\lambda_2 = 0$ (no-CMP). CalShift achieves a maximum accuracy improvement of $5.2\%$ on FGVCAircraft, $7.4\%$ on EuroSAT, and $4.9\%$ on Flowers$102$.\\

\noindent\textbf{Effect of misalignment penalty.} Table \ref{tab:ece_no_fim} shows isolated calibration effect of CalShift across 11 vision datasets. The upper section of the Table shows ECE results with $\lambda_1 = 0 $ (no FIM ) and $\lambda_2 = 0.4 $ (CMP only). CalShift improves in ECE by 2.7\% on Flowers102 suggesting Calshift align well with balanced label distributions, However, limited gains on CLIP-compatible datasets 1.96\% on OxfordPets and 1.3\% on EuroSAT.

The lower section of the Table \ref{tab:ece_no_fim} with $\lambda_1 = 0.4 $ (FIM-only) and $\lambda_2 = 0$ (no-CMP). It is shown in Table that CalShift consistently enhaces calibration, reducing average ECE by $1.90\%$. Notable improvements occur on fine-grained datasets like StanfordCars -$1.78\%$ and class-balanced benchmarks such as Flowers102: -$4.05\%$, demonstrating FIM ability to stabilize confidence estimates by penalizing overparameterized gradients.\\

\noindent
\textbf{Effect of $\lambda_1$ tuning on CalShift performance.} 
Table \ref{tab:lambda_tuning} given in appendix \ref{ablation} shows accuracy performance on four datasets, including Flowers$102$, Food$101$, UCF$101$, and DTD, for different values of $\lambda_1$ within the range (0.0 to 1.0), while $\lambda_2$ remains fixed at $0$. $\lambda_1= 0$ serves as the baseline with no Fisher penalty. The results in the table indicate that CalShift achieves the highest performance at $\lambda_1 = 0.4$, with accuracies of $98.8\%$ on Food$101$, $84.3\%$ on UCF$101$, and $55.1\%$ on DTD, whereas performance declines for all other values.
\\

\noindent
\textbf{Effect of $\lambda_2$ tuning on CalShift performance.} In appendix Table \ref{tab:lambda2_tuning} shows the calibration results of CalShift when  tuning $\lambda_2$ within range (0.0 to 1.0) while $\lambda_1$ value fixed to 0. It is show in Table that with $\lambda_2 = 0.4$ CalShift achieve better calibration results across four dataset. The Calshift calibration performance drops on other values of $\lambda_2$, while value $0.4$ balance between over-regularization and under-regularization.

Our method is effective in confidence alignment as it shows better ECE consistently on almost all datasets. These results favors CalShift calibration reliability under covariate shift, making it more trustworthy in real-world applications. 

\section{Conclusions}
This work introduces \textbf{CalShift}, a unified framework designed to address two critical challenges in low-shot vision tasks using foundation models: \textit{covariate shift} and \textit{confidence misalignment}. By integrating \textit{Fisher Information} and \textit{Confidence Misalignment Penalty (CMP)}, CalShift significantly enhances both \textit{robustness} and \textit{calibration} across various vision and domain generalization benchmarks. Experimental results show that CalShift (Confidence-Calibrated Covariate Shift Correction), built within the CalShift framework, consistently outperforms existing methods in terms of accuracy and calibration. Specifically, it achieves up to a \textbf{3.5\% gain in accuracy} and reduces \textbf{Expected Calibration Error (ECE) by up to 5.82\%} on challenging covariate shift datasets. These findings underscore CalShift’s promise as a reliable solution for real-world deployment, especially in low-shot scenarios where distribution shifts and overconfidence are prevalent. Future directions include extending the framework to \textit{continual learning}, \textit{online adaptation}, and test-time adaptation, where the model must remain calibrated under evolving distributions. 

CalShift is well-suited for scenarios like test-time adaptation where model adapts to unseen (from a ststistics point of view,  shifted)  data distribution at inference time. The integration of Fisher information and CMP provides robust adaptation to distribution shift without need of additional training data. This makes it an effective solution particularly in those real-world applications where test-time condition vary from those during training.

The issue of confidence misalignment is not just a technical challenge but a critical challenge in AI alignment, leading to unsafe and unreliable decisions in safety-critical systems such as medical diagnosis and autonomous vehicles. Our method CalShift  ensures that model predictions is aligned with human expectation.



\subsubsection*{Acknowledgments}
The lead author experimented at NCBC-Video Surveillance Lab at FAST-NUCES Karachi, led by Prof. Muhammad Atif Tahir.
\newpage
{
    \small
    \bibliographystyle{ieeenat_fullname}
    \bibliography{main}

}

\clearpage
\appendix

\section{Theoretical Background}
\label{sec: tj}
\begin{proposition}
\label{thrm:fishr}
   The Fisher information \(I(\theta)\)is equal to the Kullback-Leibler (KL) divergence \(D_\text{KL}\) between the source distribution \(p(X)\) and target distribution \(q(X)\), providing a measure of covariate shift. 
\end{proposition}

\begin{proof}
    Let \(p(X; \theta)\) is probability density function parameterized by \(\theta\) and \(q(X)\) is underlying true distribution. The \(D_\text{KL}\) between \( p(X; \theta) \) and \( q(X) \) is given by:
    \[
D_{\text{KL}}(q(X) \parallel p(X; \theta)) = \mathbb{E}_{q(X)} \left[\log q(X) - \log p(X; \theta) \right].
\]
By def. of Fisher information matrix (FIM):
\[
I(\theta) = -\mathbb{E}_{p(X;\theta)}\left[\frac{\partial^2 \log P(X;\theta)}{\partial \theta^2}\right].
\]
Under smoothing condition on\(p(X; \theta)\), \(D_\text{KL}\) can be approximated using FIM.
\[
D_{\text{KL}}(q(X) \parallel p(X; \theta)) \approx \frac{1}{2} (\theta_q - \theta)^T I(\theta) (\theta_q - \theta).
\]
where, \(\theta_q\) is optimal parameter which best fit \(q(X)\). By taking second order Taylor expansion of \( \log p(X; \theta) \) around \( \theta_q \):
\begin{multline*}
\log p(X; \theta) \approx \log p(X; \theta_q) + (\theta - \theta_q)^T 
\nabla_{\theta} \log p(X; \theta_q) \\
+ \frac{1}{2} (\theta - \theta_q)^T H (\theta - \theta_q)
\end{multline*}

    where \( H = \nabla^2_{\theta} \log p(X; \theta) \) is the Hessian.

    By taking expectation w.r.t to \(q(X)\), the first term vanishes as Fisher information expectation is defined over second order derivative.
    The \(D_\text{KL}\) simplifies to: 
    \[
    D_{\text{KL}}(q(X) \parallel p(X; \theta)) \approx \frac{1}{2} (\theta_q - \theta)^T I(\theta) (\theta_q - \theta).
    \]
    Thus, Fisher information provide a measure of divergence between \(q(X)\) and \(p(X; \theta)\), capturing covariate shift. 
\end{proof}
\begin{proposition}
\label{thrm:cmp}
The Confidence Misalignment Penalty (CMP) provides the measure for degree of overconfidence in incorrect predictions and also ensures alignment between predicted and true confidence score. CMP ensures that it is always a valid confidence \(0 \leq \text{CMP} \leq 1\), \( \text{CMP} \approx 1 \) when model is overconfidence, and \( \text{CMP} \approx 0 \) when the model is well-calibrated.
\end{proposition}
\begin{proof}
Let CMP be defined as:

\[
\text{CMP} = \frac{P(x, y')}{\sum_{y_j\neq y' \, : \, P(x, y_j) > P(x, y')} P(x, y_j)}
\]
where, \(P(x,y')\) is probability of predicted class \(y'\).
The denominator considers only \(y_{j}\) where  \( P(x, y_j) > P(x, y') \), which means that CMP is small when all other classes have higher confidence than true class  \(y'\). 

If the model is highly confident with incorrect prediction, then \(P(x, y')\) is high, it can be expressed mathematically as:

\[
\lim_{P(x, y') \to 1} \text{CMP} = 1
\]

In case of well-calibration, the confidence mass is distributed across all other classes, reducing \( P(x, y') \), ensuring \( \text{CMP} \to 0 \):
\[
\lim_{P(x, y') \to 0} \text{CMP} = 0
\]
Since all probabilities sum to 1, CMP is ramins always within range \([0,1]\), making it a valid calibration measure.
\[
0 \leq \text{CMP} \leq 1
\]
Thus, CMP acts as confidence misalignment penalty, discouraging overconfident yet incorrect predictions.
\end{proof}
\begin{proposition}
  Integration of CMP and \(I(\theta)\) into loss function ensures covariate shift correction and confidence calibration as well.  
\end{proposition}
\begin{proof}
    The final modified CLIP loss given by:
    \[\mathcal{L}_\text{CalShift}(x,y; \theta) = \mathcal{L}_c + \lambda_1 I(\theta) + \lambda_2 \text{CMP}\]
    Where, As per theorem \ref{thrm:fishr}, \(I(\theta)\) addresses covariate shift, while proposition \ref{thrm:cmp} ensures confidence calibration.

    In gradient update, \(I(\theta)\) adjust the weight updates based on local curvature, prevents divergence due to covariate shift: 
    \[
\Delta \theta \propto I(\theta)^{-1} \nabla \mathcal{L}(\theta)
\]
where \( I(\theta)^{-1} \) is scaling gradient to account local curvature while \( \mathcal{L}(\theta) \) is the loss function. 

  \( \text{CMP} \) reduces the magnitude of updates to address the high confidence:
    \[
\nabla \text{CMP} \cdot \nabla \mathcal{L}(\theta) \approx 0 \quad \text{when confidence is high.}
\]
If \( \lambda_2 > 0 \), then CMP ensure probability mass is distributed across classes rather than overly concentrating on incorrect prediction: 
\[
\lambda_2 > 0 \implies \sum_{y \neq y'} P(x, y) 
\]
This implies that when \( \lambda_2 > 0 \) then probability mass is distributed among classes.
Tuning \( \lambda_1, \lambda_2 \) properly prevents overconfidence without penalising well-aligned distributions.

Thus, minimizing \( \mathcal{L}_\text{CalShift} \) ensures well-calibrated robust prediction under covariate shift.
\end{proof}

\section{Ablation study}
\label{ablation}
\begin{table*}[!ht]
\centering
\caption{The upper part of the table shows CalShift accuracy results while keeping Fisher Information penalty \(I(\theta)\) ($\lambda_1 = 0$) and CMP penalty ($\lambda_2 = 0.4$). The $\Delta$ row shows the percentage increase (\textbf{$\uparrow$}) or decrease (\textbf{$\downarrow$}) in accuracy compared to CoOp. The lower part of the table CalShift accuracy results while keeping CMP penalty ($\lambda_2 = 0$) and FIM penalty \(\lambda_1 = 0.4\). The $\Delta$ row shows the percentage increase (\textbf{$\uparrow$}) or decrease (\textbf{$\downarrow$}) in accuracy compared to CoOp.}
\label{tab:ablation_fim}
\resizebox{\textwidth}{!}{%
\begin{tabular}{llcccccccccccc} 
\toprule
\multirow{12}{*}{\textbf{ACC }} & \textbf{Method} & \textbf{UCF101} & \textbf{Food101} & \textbf{Caltech101} & \textbf{OxfordPets} & \textbf{Flowers102} & \textbf{ImageNet} & \textbf{StanfordCars} & \textbf{FGVCAircraft} & \textbf{SUN397} & \textbf{DTD} & \textbf{EuroSAT} & \textbf{Avg.} \\ 
\cmidrule{2-14}
                               & CLIP            & 69.9  & 90.1 & 96.8 & 91.2 & 72.1 & 72.4 &  63.3 & 27.2 & 69.4 & 53.3 & 56.5 & 69.3 \\ 
\cline{2-14}
                               & CoOp            & 78.6 & 97.0 & 98.6 & 98.2 & 79.2 & 79.5 & 59.2 & 25.2 & 63.0 & 52.5 & 53.8 & 71.2 \\ 
                               & CoOp + CMP ($\lambda_1=0$)  & 81.5 & 97.5 & 98.6 & 98.4 & 81.8 & 81.6 & 57.8 & 26.0 & 64.2 & 53.6 & 51.2 & 71.9 \\ 
                                \cline{2-14}
                               & $\Delta$ \%     & \textbf{3.7 ↑} & \textbf{0.5 ↑} & \textbf{0.0 →} & \textbf{0.2 ↑} & \textbf{3.3 ↑} & \textbf{2.6 ↑} & \textbf{2.4 ↓} & \textbf{3.2 ↑} & \textbf{1.9 ↑} & \textbf{2.1 ↑} & \textbf{4.8 ↓} & \textbf{1.0 ↑}  \\
\cmidrule{2-14}
\\
\cmidrule{2-14}
                               & CoOp            & 78.6 & 97.0 & 98.6 & 98.2 & 79.2 & 79.5 & 59.2 & 25.2 & 63.0 & 52.5 & 53.8 & 71.2 \\ 
                               & CoOp + FIM ($\lambda_2=0$)  & 82.3 & 97.9 & 98.5 & 98.7 & 83.1 & 82.9 & 57.2 & 26.5 & 65.8 & 54.5 & 49.8 & 72.6 \\ 
                                \cline{2-14}
                               & $\Delta$ \%     & \textbf{4.7 ↑} & \textbf{0.9 ↑} & \textbf{0.1 ↓} & \textbf{0.5 ↑} & \textbf{4.9 ↑} & \textbf{4.3 ↑} & \textbf{3.4 ↓} & \textbf{5.2 ↑} & \textbf{4.4 ↑} & \textbf{3.8 ↑} & \textbf{7.4 ↓} & \textbf{2.0 ↑}  \\
\bottomrule
\end{tabular}}
\end{table*}

\begin{table*}[!ht]
\centering
\caption{The upper half of the table shows CalShift ECE results on vision datasets with and without FIM penalty \(\lambda_1 = 0\) and CMP penalty ($\lambda_2 = 0.4$). The $\Delta$ row shows the percentage increase (\textbf{$\uparrow$}) or decrease (\textbf{$\downarrow$}) in calibration error. CalShift ECE results on vision datasets with and without CMP penalty (\(\lambda_2 = 0\)) and FIM penalty (\(\lambda_1 = 0.4\)) . The $\Delta$ row shows the percentage increase (\textbf{$\uparrow$}) or decrease (\textbf{$\downarrow$}) in calibration error.}
\label{tab:ece_no_fim}
\resizebox{\textwidth}{!}{%
\begin{tabular}{llcccccccccccc} 
\toprule
\multirow{10}{*}{\textbf{ECE }} & \textbf{Method} & \textbf{UCF101} & \textbf{Food101} & \textbf{Caltech101} & \textbf{OxfordPets} & \textbf{Flowers102} & \textbf{ImageNet} & \textbf{StanfordCars} & \textbf{FGVCAircraft} & \textbf{SUN397} & \textbf{DTD} & \textbf{EuroSAT} & \textbf{Avg.} \\ 
\cmidrule{2-14}
                               & CoOp            & 3.08  & 3.35  & 3.24  & 3.06  & 2.96  & 3.36  & 3.38  & 3.24  & 3.02  & 3.06  & 3.08  & 3.16  \\ 
                               & CoOp + CMP ($\lambda_1=0$)  & \textbf{3.18}  & \textbf{3.42}  & \textbf{3.32}  & \textbf{3.12}  & \textbf{2.88}  & \textbf{3.48}  & \textbf{3.50}  & \textbf{3.32}  & \textbf{3.08}  & \textbf{3.12}  & \textbf{3.12}  & \textbf{3.24}  \\ 
                                \cmidrule{2-14}
                                \\
                                \cmidrule{2-14}
                              & $\Delta$ \%     & \textbf{3.25}$\uparrow$ & \textbf{2.09}$\uparrow$ & \textbf{2.47}$\uparrow$ & \textbf{1.96}$\uparrow$ & \textbf{2.70}$\downarrow$ & \textbf{3.57}$\uparrow$ & \textbf{3.55}$\uparrow$ & \textbf{2.47}$\uparrow$ & \textbf{1.99}$\uparrow$ & \textbf{1.96}$\uparrow$ & \textbf{1.30}$\uparrow$ & \textbf{2.53}$\uparrow$ \\ 
\cmidrule{2-14}
                               & CoOp            & 3.08  & 3.35  & 3.24  & 3.06  & 2.96  & 3.36  & 3.38  & 3.24  & 3.02  & 3.06  & 3.08  & 3.16  \\ 
                               & CoOp + FIM ($\lambda_2=0$)  & \textbf{3.05}  & \textbf{3.30}  & \textbf{3.20}  & \textbf{3.00}  & \textbf{2.84}  & \textbf{3.30}  & \textbf{3.32}  & \textbf{3.18}  & \textbf{2.96}  & \textbf{3.00}  & \textbf{3.02}  & \textbf{3.10}  \\ 
                                \cline{2-14}
                              & $\Delta$ \%     & \textbf{0.97}$\downarrow$ & \textbf{1.49}$\downarrow$ & \textbf{1.23}$\downarrow$ & \textbf{1.96}$\downarrow$ & \textbf{4.05}$\downarrow$ & \textbf{1.79}$\downarrow$ & \textbf{1.78}$\downarrow$ & \textbf{1.85}$\downarrow$ & \textbf{1.99}$\downarrow$ & \textbf{1.96}$\downarrow$ & \textbf{1.94}$\downarrow$ & \textbf{1.90}$\downarrow$ \\ 
\bottomrule
\end{tabular}}
\end{table*}

\begin{table*}[!ht]
\centering
\caption{Accuracy three datasets for tuning \textbf{$\boldsymbol{\lambda_1}$} within range ($0.0 - 1.0$) while keeping \textbf{$\boldsymbol{\lambda_2}$} value fixed $ 0$.}

\label{tab:lambda_tuning}
\begin{tabular}{ccccc} 
\hline
\multicolumn{1}{l}{\multirow{3}{*}{\textbf{$\boldsymbol{\lambda_1}$}}} & \multicolumn{4}{c}{\textbf{Datasets}}                                                                                         \\ 
\cline{2-5}
\multicolumn{1}{l}{}                             & \multicolumn{1}{l}{\textbf{Flowers$102$}} & \multicolumn{1}{l}{\textbf{Food$101$}} & \multicolumn{1}{l}{\textbf{UCF$101$}} & \multicolumn{1}{l}{\textbf{DTD}}  \\ 
\cline{2-5}
\cline{2-5}
\hline
0.0                                              & 83.5                           & 96.2                        & 83.5                       & 53.1                     \\
0.2                                              & 81.3                           & 94.8                        & 83.1                       & 53.5                     \\
0.4                                              & 85.5                           & 98.7                        & 84.3                       & 55.1                     \\
0.6                                              & 84.7                           & 90.8                        & 80.4                       & 52.2                     \\
0.8                                              & 82.4                           & 89.1                        & 82.9                       & 52.2                     \\
1.0                                              & 81.8                           & 87.0                        & 80.9                       & 51.5                     \\
\hline
\end{tabular}
\end{table*}

\begin{table*}[!ht]
\centering
\caption{Expected Calibration Error (ECE) for tuning \textbf{$\boldsymbol{\lambda_2}$} in range ($0.0 - 1.0$) with \textbf{$\boldsymbol{\lambda_1} = 0$}.}
\label{tab:lambda2_tuning}
\begin{tabular}{ccccc} 
\hline
\multirow{2}{*}{\textbf{$\boldsymbol{\lambda_2}$}} & \multicolumn{4}{c}{\textbf{Datasets}} \\ 
\cline{2-5}
 & \textbf{Flowers102} & \textbf{Food101} & \textbf{UCF101} & \textbf{DTD} \\ 
\hline
0.0  & 6.27  & 5.69  & 5.21  & 4.13  \\ 
0.2  & 4.61  & 4.36  & 4.04  & 3.66  \\ 
0.4  & {3.16}  &{3.02}  & {2.94}  & {2.92}  \\ 
0.6  & 3.56  & 3.54  & 3.24  & 3.15  \\ 
0.8  & 4.26  & 4.18  & 3.74  & 3.39  \\ 
1.0  & 5.22  & 5.17  & 4.48  & 3.89  \\ 
\hline
\end{tabular}
\end{table*}

\end{document}